\documentclass[pdflatex,sn-mathphys-num]{sn-jnl}

\usepackage{booktabs}
\usepackage{multirow}
\usepackage{tabularx}
\usepackage{adjustbox}
\usepackage{array}
\usepackage{graphicx}%
\usepackage{multirow}%
\usepackage{amsmath,amssymb,amsfonts}%
\usepackage{amsthm}%
\usepackage{mathrsfs}%
\usepackage{textcomp}
\usepackage[title]{appendix}%
\usepackage{xcolor}%
\usepackage{soul}%
\usepackage{wasysym}
\usepackage{multicol}
\usepackage{textcomp}%
\usepackage{manyfoot}%
\usepackage{booktabs}%
\usepackage{algorithm}%
\usepackage{algorithmicx}%
\usepackage{algpseudocode}%
\usepackage{listings}%
\usepackage{hyperref}
\usepackage{xr}
\theoremstyle{thmstyleone}%
\theoremstyle{thmstyletwo}%

\theoremstyle{thmstylethree}%

\usepackage{xcolor}
\definecolor{lightgray}{rgb}{0.92, 0.92, 0.92}
\definecolor{green}{rgb}{0.0, 0.62, 0.38}
\definecolor{red}{rgb}{0.84, 0.00, 0.25}

\begin{document}

\title{SunLightBots: Turning sunlight into self-sustained, high-power motion in soft robots}

\author[1]{\fnm{Yoo Jin Lee}}

\author[1]{\fnm{Melanie Torres Crespo}}

\author[3]{\fnm{Zhongxiao Deng}}

\author[1]{\fnm{Andrew Tejada-Vega}}

\author[1,2]{\fnm{Rebecca Kramer-Bottiglio}}

\author*[1,3]{\fnm{Wenzhong Yan}}\email{wyan@ucdavis.edu}

\author*[1]{\fnm{Yimin Luo}}\email{yimin.luo@yale.edu}

\affil[1]{\orgdiv{Mechanical Engineering}, \orgname{Yale University}, \orgaddress{\street{9 Hillhouse Avenue}, \city{New Haven}, \postcode{06511}, \state{CT}, \country{USA}}}

\affil[2]{\orgdiv{Present Address: Mechanical and Aerospace Engineering}, \orgname{Princeton University}, \orgaddress{\city{Princeton}, \postcode{08540}, \state{NJ}, \country{USA}}}

\affil[3]{\orgdiv{Mechanical and Aerospace Engineering}, \orgname{University of California, Davis}, \orgaddress{\street{One Shields Avenue}, \city{Davis}, \postcode{95616}, \state{CA}, \country{USA}}}


\abstract{

Soft robots built from active materials can directly convert environmental stimuli into motion, offering a route to energy autonomy and embodied control. However, environmental energy is generally harvested at low power, whereas rapid, mechanically demanding actions require high mechanical output power. Reconciling this mismatch while sustaining self-regulated cyclic actuation is challenging, particularly at the subgram scale, where limited payload capacity constrains the integration of conventional power and control hardware.
Here, we present \emph{SunLightBots}, a class of subgram, sunlight-powered, untethered soft robots that achieve autonomous locomotion under continuous illumination near 1 Sun (100~mW/cm\textsuperscript{2}). Their operation is enabled by a self-resetting snapping (SRS) mechanism that integrates liquid crystal elastomers (LCEs) with snap-through instability to accumulate solar energy supplied at low power, release it as rapid, high-power motion, and reset through photothermal--mechanical feedback.
Through coordinated optimization of the LCE actuation temperature, photothermal conversion, actuation stress, and structural energy barrier, the system operates near 1 Sun and achieves a peak mass-specific mechanical power of 89~W/kg. At this irradiance, the optimized LCE generates an actuation stress of 0.1~MPa at approximately 60~${}^{\circ}$C. \emph{SunLightBots} demonstrate sustained swimming, crawling, and repeated impulsive jumping. Under 1.5 Suns, the swimmer advances more than three body lengths per cycle. This work provides a materials--mechanics route toward autonomous soft robots powered entirely by the environment.}

\keywords{liquid crystal elastomer, microrobot, soft robot, instability, sunlight-powered}

\maketitle

\section*{Introduction}\label{sec:intro}

Soft robots expand robotic capabilities through their intrinsic structural compliance and access to a broad palette of functional materials~\cite{rus2015design}. Active materials are especially attractive because they can convert environmental stimuli into deformation, potentially integrating energy harvesting, actuation, and physical regulation within the robot body~\cite{sitti2021physical}. However, achieving energy autonomy together with rapid, mechanically demanding actions remains challenging. Functions such as rapid shape change, jumping, and effective propulsion require mechanical work to be delivered over a short duration and therefore demand high mechanical output power~\cite{yan2023origami}. Such capabilities commonly rely on batteries, high-voltage electronics, pressurized fluids, or external power \cite{jung2024untethered,rich2018untethered}, whereas environmentally powered systems generally produce one-shot, lower-power motion. This tradeoff is particularly severe at the subgram scale, where limited payload capacity constrains the integration of conventional power and control hardware~\cite{bergbreiter2007design}.

Natural systems demonstrate how energy autonomy can coexist with high-power motion. Organisms acquire energy from their surroundings and regulate its conversion, storage, and use within the body. The Venus flytrap is a prime example of an organism that derives energy from sunlight through photosynthesis and uses gradual active deformation to store elastic energy in its leaves; snap-through instability then releases this energy rapidly to close the trap and capture prey~\cite{forterre2005venus}. Jumping insects use muscle to load elastic structures before releasing the stored energy impulsively~\cite{ilton2018principles}. These mechanisms do not increase the available energy; instead, they concentrate its release in time and thereby amplify mechanical power. This integration of energy acquisition, actuation, energy storage, power amplification, and regulation provides a model for autonomous soft robots powered by low-power environmental energy.

Sunlight is a particularly attractive energy source for such systems because it is abundant, renewable, and can potentially be harvested directly by the robot body~\cite{jafferis2019untethered,shen2024sunlight}. Photoresponsive liquid crystal elastomers (LCEs) provide a promising route for direct photomechanical conversion \cite{camacho2004fast,ceamanos2020four} because their transition temperature \cite{yao2023enabling,evenchik2026reducing}, actuation strain \cite{yakacki2015tailorable}, and deformation direction \cite{kotikian20183d,sydney2016biomimetic} can be tailored through material synthesis and programming. LCEs and related responsive materials have been used to generate self-sustained oscillation, locomotion, gripping, and object manipulation~\cite{zhao2025antagonistic,zhao2022twisting,kumar2016chaotic,gelebart2017mastering,zhai2023self,li2021light,hu2021phototunable}. Physical feedback mechanisms such as self-shadowing allow constant illumination to provide both energy and motion regulation without electronic control. Recent advances have reduced the threshold irradiance for self-sustained oscillation to approximately 1 Sun; however, robotic functions such as walking, sailing, and flapping have still required irradiances of approximately 2 Suns or higher~\cite{zhao2023sunlight}. Light-powered self-sustained jumping has also been demonstrated using snap-through instability and twisted-LCE architectures~\cite{yan2025self,qi2025self}, but these systems require irradiances exceeding 6 Suns. To our knowledge, autonomous, cyclic, and high-power mechanical output under unconcentrated sunlight has not yet been demonstrated.

To address this challenge, we develop a self-resetting snapping (SRS) mechanism that integrates solar-energy transduction with mechanical energy storage and power amplification. An active LCE--elastomer composite combines a low actuation temperature, efficient photothermal conversion, and high actuation stress and strain to slowly load a monostable structure (with snap-through). Its energy barrier is tuned to store substantial energy while remaining accessible under weak illumination. Snap-through rapidly releases the stored energy, while coupled photothermal--mechanical feedback and environmental cooling reset the structure, enabling repeated load--release cycles without external sensors, controllers, or triggers.

\begin{figure}
\centering
\includegraphics[width=\textwidth]{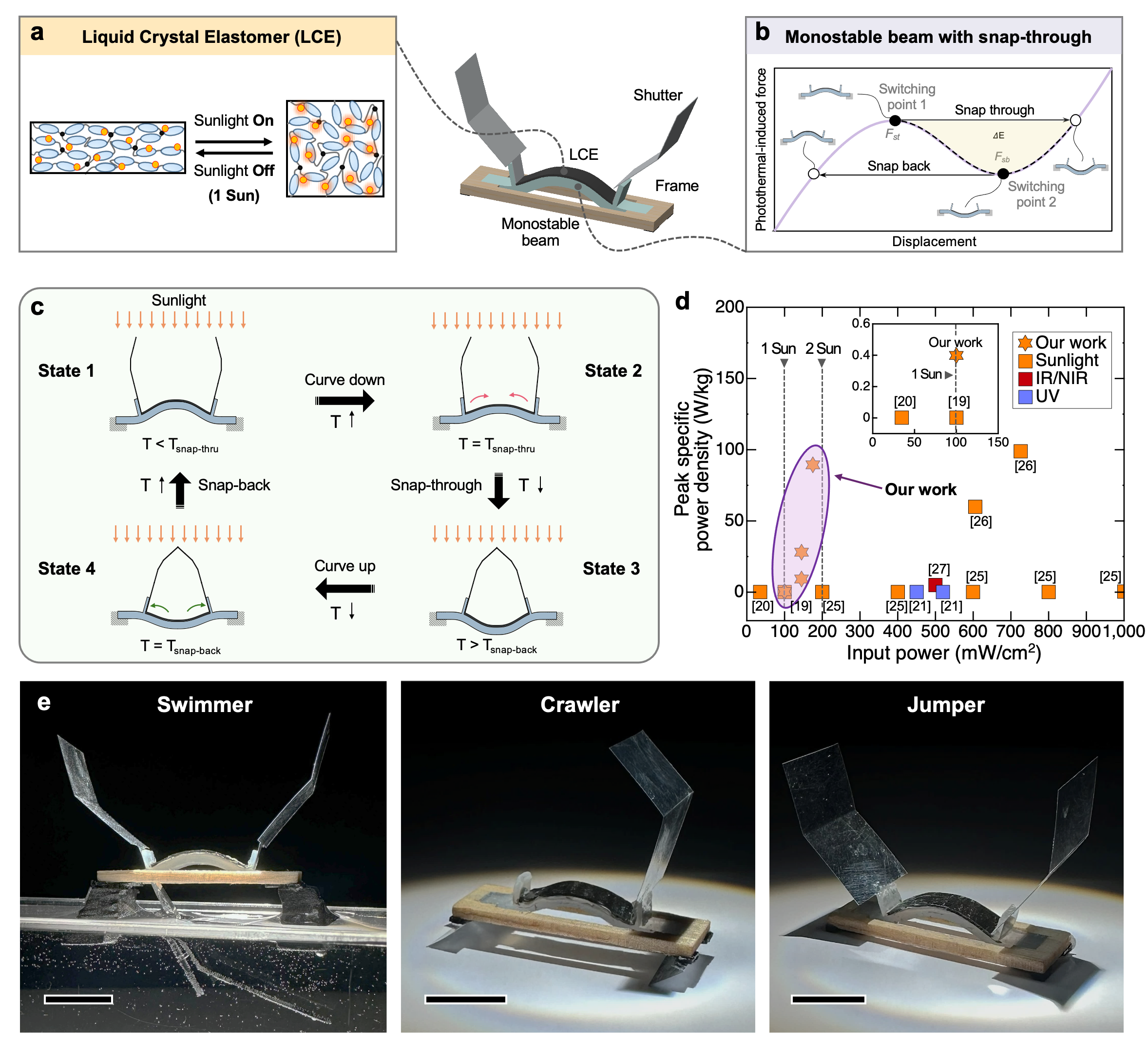}
   \caption{\textbf{Overview of the SunLightBots.} (\textbf{a}) Photothermal actuation of liquid crystal elastomers (LCEs), with the light irradiance required for actuation reduced to 1 Sun illumination (100 mW cm$^{-2}$).  (\textbf{b}) Schematic force–displacement diagram of a monostable LCE–curved beam composite. As the (photo)thermally generated LCE force along the contraction direction reaches the snap-through threshold $F_{st}$, the beam snaps downward, abruptly releasing energy $\Delta E$ (shaded in green). As the force falls below the snap-back threshold $F_{sb}$, the beam returns to its upward configuration. (\textbf{c}) Coupling low-energy photothermal actuation of LCEs with a snap-through mechanism, the device enables reversible cycles of high-power mechanical output from a low-power solar input.  (\textbf{d}) Ashby plot comparing the peak output specific power density of previously reported photothermally driven actuator systems. Our work is indicated by stars, each representing the optimized LCE formulation coupled to a different elastomer. The inset shows a magnified view of the data points near 1 Sun.  (\textbf{e}) SRS-based SunLightBots can achieve various locomotion, including swimming, crawling, and jumping. The scale bars are 1 cm. 
}
\label{fig:overview}
\end{figure}

We have implemented these strategies in \emph{SunLightBots}, a family of untethered soft microrobots, operating at irradiances near 1 Sun (100~mW/cm\textsuperscript{2}, Fig.~\ref{fig:overview}). At this irradiance, the LCE generates an actuation stress of $0.1~\mathrm{MPa}$ at approximately $60~^\circ\mathrm{C}$, and the SRS mechanism achieves a peak mass-specific mechanical power of 89~W/kg, which compares favorably with values reported for previous photothermally driven actuator systems (Fig.~\ref{fig:overview}d and Supplementary Section S1 for the detailed calculation).
The same mechanism drives sustained swimming, crawling, and repeated impulsive jumping, demonstrating a general strategy for converting slowly harvested environmental energy into rapid and mechanically effective robotic motion (Fig.~\ref{fig:overview}e). More broadly, this materials--mechanics approach provides a route toward autonomous soft robots with energy harvesting, actuation, and power amplification, to achieve cyclic locomotion under unconcentrated sunlight. 

\section*{Results}\label{sec:results}

\subsection*{Design and development of SunLightBots}

To obtain high power output from low-power input, several strategies have been widely adopted, including: 1) temporal power amplification through energy accumulation followed by rapid release~\cite{ilton2018principles}, 2) mechanical amplification of force or displacement through compliant mechanisms or transmission systems~\cite{hu2018small}, and 3) instability-triggered energy release through snap-through or buckling~\cite{yan2025self,overvelde_amplifying_2015,lee2010first}. Taking inspiration from these strategies, we combine an elastomer beam with a photoresponsive LCE to form a monostable composite with snap-through (Fig.~\ref{fig:overview}a and b). The LCE actuator converts low-intensity light energy into mechanical contraction, activating the composite beam and triggering a rapid release after it overcomes the energy barrier. The shutters serve to control the exposure of the LCE to illumination, which, when combined with the monostable composite, enables cyclic snapping and therefore untethered, sustained autonomous cyclic actuation under continuous illumination. The shutter geometry is critical to autonomous cycling. To accomplish this, it must (i) maximize LCE exposure before snap-through and (ii) fully shade the LCE after snap-through to enable cooling and snap-back (see Materials and Methods for the detailed design).

The cyclic snapping derives from a self-shadowing-induced built-in feedback loop (Fig.~\ref{fig:overview}c) based on the dynamic light–material interplay~\cite{qian2019artificial, zhao_soft_2019,hua2021swaying, yan2021artificial,he2012synthetic}. Specifically, light irradiation photothermally heats the LCE (State 1), and the heating induces LCE contraction (State 2), which triggers snapping (State 3). After snap-through, the beam configuration closes the shutters, blocking light to allow the LCE to cool and relax, thereby reopening the shutters through a snap-back (State 4). This interplay leads to cyclic snapping, enabling continuous, repeated, rapid motion. The setup presented in our work achieves a high peak specific power output relative to input power amongst all light-driven machines (Fig. \ref{fig:overview}d), with particularly strong performance under low-energy illumination conditions. Detailed peak specific power output definitions, calculations, and literature data are provided in Supplementary Section S1 and Supplementary Table S1.

Key considerations for enabling cyclic snap-through under low-intensity illumination include the LCE thermomechanical stress, photothermal conversion efficiency, and the energy barrier of the monostable composite. The LCE must absorb sufficient light and generate enough stress at the resulting temperature to trigger snap-through. Meanwhile, the energy barrier must be carefully tuned to permit triggering under weak illumination while retaining sufficient energy storage for desired motions. We therefore developed a low-transition-temperature, high-stress LCE and matched it with the energy barrier of the monostable elastomer. Using a targeted search of the material formulation space, we reduced the required input irradiance to nearly 1 Sun while maintaining cyclic, high-power actuation for self-sustained robots, as detailed in the following sections.


\begin{figure}
\centering
\includegraphics[width=\textwidth]{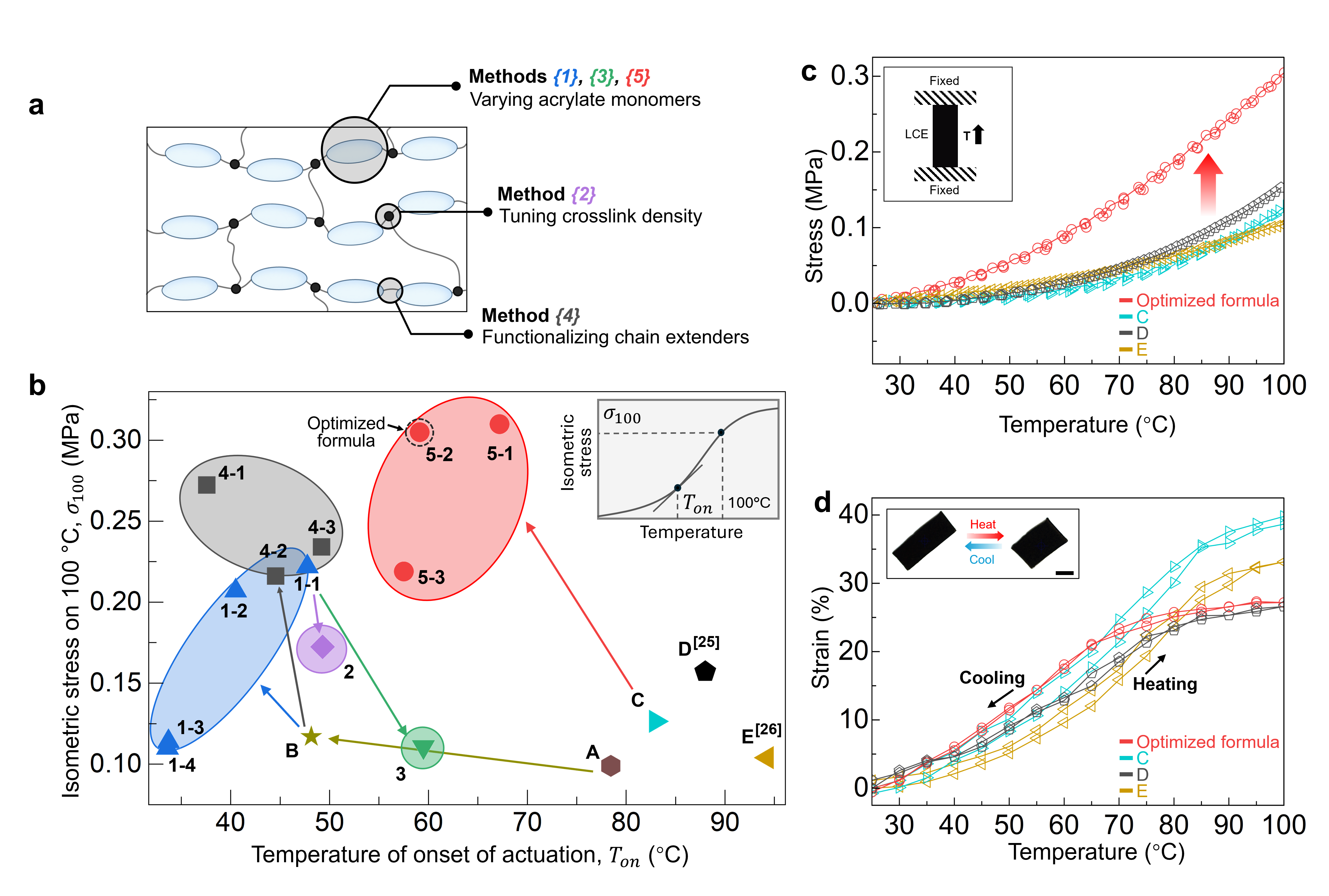}
   \caption{\textbf{Optimization of the thermomechanical properties of LCEs.} (\textbf{a}) Schematic of strategies used to develop LCEs that actuate under low energy input and produce high mechanical output. Methods 1, 3, and 5 vary the types and weight fractions of liquid-crystalline and non-liquid-crystalline acrylate monomers. Method 2 varies the crosslink density, and Method 4 varies the chain extender type. Details of the compositions are presented in Supplementary Table S2. (\textbf{b}) Comparison of energy input for actuation and mechanical output of LCEs developed using the applied strategies. Formulations A to E are either unmodified or taken from the literature, whereas Methods 1 to 5 describe the strategies used to modify them, as shown in Panel (a). The numerical sublabels then identify formulations obtained using each method. Energy input and mechanical output are evaluated by the temperature of onset of actuation and isometric stress at 100 °C, respectively. LCEs with the optimized formulation actuate around 60 °C and generate actuation stress of approximately 0.3 MPa. (\textbf{c})  Isometric stress as a function of temperature for the optimized formulation and controls C and D.
   (\textbf{d})   Actuation strain as a function of temperature for the optimized formulation and controls. Red $\Circle$'s denote the optimized formulations, gold $\triangleleft$'s denote Formulation D, and cyan $\triangleright$'s denote Formulation C, as indicated in Panel (\textbf{b}).}
\label{fig:material_optimization}
\end{figure}

\subsection*{Optimizing thermomechanical performance of LCE}

We systematically optimize material formulations to develop LCEs capable of generating sufficient mechanical output under low-power input. While LCEs are widely studied for their large actuation strains, sufficient actuation stress is critical for translating the deformation into useful mechanical work for robotic applications. In our device, the LCE must generate sufficient force within a limited mass budget to drive the monostable structure over its mechanical energy barrier and trigger snap-through. The actuation temperature imposes an additional constraint. A threshold too close to ambient temperature can increase susceptibility to environmental temperature fluctuations, whereas variations in convective heat exchange can affect whether and when this threshold is reached, compromising actuation control. Conversely, an excessively high threshold increases the heating required and can limit operation under 1 Sun irradiation. Therefore, the design objective here is to achieve high actuation stress with an actuation onset, defined as the temperature at which measurable thermally induced deformation begins, near 50–60 °C. This provides a margin above ambient conditions while remaining accessible under solar heating. 

Meeting these requirements demands coordinated control of network properties, because modifications that lower the actuation temperature can also reduce the stress available to drive snap-through.
Although several strategies for tuning actuation stress and actuation temperature have been reported \cite{guo2024hydrogen,bauman2022actuation,zhan2023cold, saed2019molecularly,barnes2022understanding}, the systematic integration and comparison of these factors across multiple formulations remain underexplored, particularly for low-energy actuation. Rather than considering these approaches individually, we systematically evaluate multiple molecular and network design strategies across two reaction chemistries, amine–acrylate and thiol–Michael systems. These strategies include variations in monomer composition, chain-extender chemistry, and crosslink density (Fig.~\ref{fig:material_optimization}a), to identify formulations that combine low-temperature actuation with high stress to drive snap-through under low-energy input. Details of the chemical structures (Supplementary Fig. S1), reaction schemes (Supplementary Figs. S2 and S3), and compositions of the LCE formulations are provided in the SI (Supplementary Table S2).

We first investigate acrylate-amine Michael addition as a reaction pathway (Fig.~\ref{fig:material_optimization}b). Initial modifications target lowering the thermal threshold for actuation. Incorporating the long-chain extender dodecylamine substantially lowered the actuation temperature (formulation A). Previous studies have shown that non-liquid-crystalline (non-LC) or hydrophilic acrylates can lower transition temperatures. We therefore introduce PEGDA to further reduce the actuation temperature (formulation B). This formulation provides a common baseline for evaluating how additional changes in network composition and intermolecular interactions could improve the balance between actuation temperature and stress.

We explore two complementary approaches: varying the LC monomers and introducing hydrogen-bond-forming chain extenders. In the first approach, we incorporate C6BAPE, an LC monomer with weaker $\pi$–$\pi$ interactions, into formulation C to further lower the actuation temperature (Method 1). Among the three formulations tested, formulation 1-1 offers the best balance between low-temperature actuation and actuation stress and is therefore selected for further optimization. We then seek to increase the actuation stress while maintaining the low actuation temperature. However, increasing either the crosslink density (Method 2) or the fraction of LC monomers relative to non-LC components (Method 3) raises the actuation temperature while reducing the generated stress, attributed to restricted polymer chain mobility and reduced network flexibility, respectively. Both modifications therefore move the material away from the desired combination of low-temperature, high-stress actuation.

In parallel, we introduce hydrogen-bond-forming chain extenders into formulation C while keeping the acrylate monomer composition unchanged (Method 4). This approach increases the actuation stress while further lowering the actuation temperature, improving both properties simultaneously. Among the amine–acrylate formulations tested, Method 4 offers the best balance between actuation temperature and stress. These results show that the trade-off between actuation temperature and stress strongly depends on how the network is modified. While increasing the LC content or crosslink density is unfavorable, introducing hydrogen-bond-forming chain extenders enables higher stress at lower actuation temperatures.

We also investigate thiol–acrylate Michael addition as an alternative chemistry to further improve the balance between actuation temperature and stress achieved with amine–acrylate LCEs. Starting from a control formulation (formulation C), we systematically vary the type and weight fraction of LC monomers to examine their effects on actuation temperature and stress (Method 5) \cite{barnes2022understanding}. Increasing the RM257 content lowers the actuation temperature while maintaining relatively high actuation stress. In particular, formulations 5-1 and 5-2 exhibit actuation onset temperatures near 60 $^\circ$C while generating higher stresses than the amine–acrylate LCEs tested in this study. Among all formulations investigated across both reaction chemistries, formulation 5-2 offers the best combination of low actuation temperature and high actuation stress, providing the mechanical output needed to drive snap-through under low-energy input. Stress and strain measurements further show that the optimized formulation generated higher actuation stress (Fig.~\ref{fig:material_optimization}c) and comparable actuation strain (Fig.~\ref{fig:material_optimization}d) over the temperature range examined when compared to three benchmark formulations: the formulation used in our previous work (formulation E) \cite{yan2025self}, the formulation reported to develop sunlight-driven soft robots (formulation D) \cite{zhao2023sunlight}, and the control (formulation A). We therefore select formulation 5-2 for subsequent device development, as it provides the combination of actuation temperature and mechanical output needed for snap-through.

\subsection*{Improving photothermal performance of LCE}

\begin{figure}
\centering
\includegraphics[width=\textwidth]{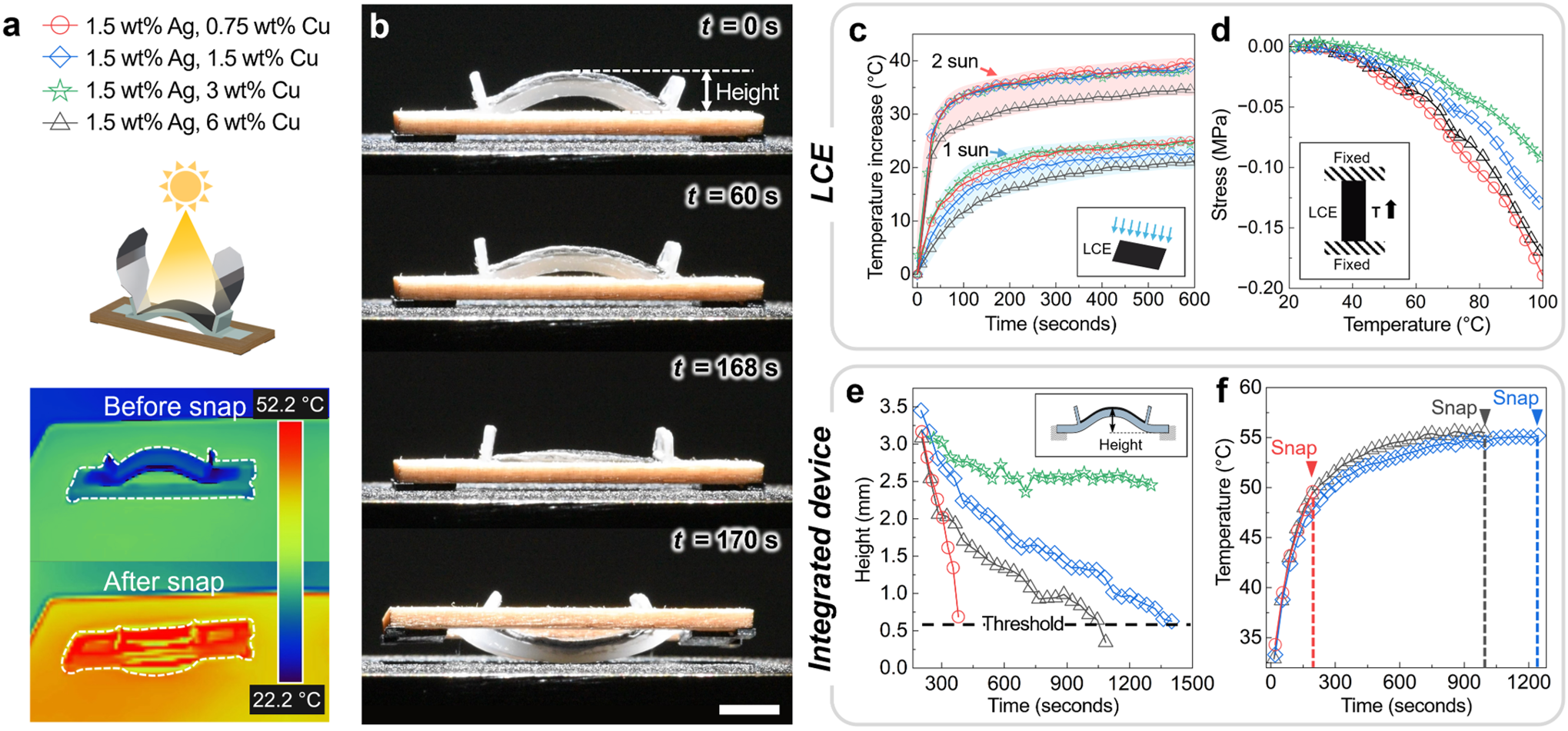}
   \caption{\textbf{Optimization of the photothermal response of LCEs and LCE-based devices.} (\textbf{a}) Schematic of photothermal actuation of a device comprising an LCE and a curved beam (top). With silver content fixed, the copper amount was varied to optimize photothermal efficiency. Device temperature throughout actuation was monitored using a thermal camera (bottom). (\textbf{b}) Representative images of light-driven actuation. LCE contraction under illumination deforms the curved beam and triggers snap-through. The scale bar is 5 mm.  (\textbf{c}) Photothermal and (\textbf{d}) mechanical responses of LCEs with different copper contents under 1 and 2 Sun illumination. (\textbf{e}) Time-dependent changes in beam height and (\textbf{f}) temperature rise of the device under 1 Sun illumination. Colors and symbols denote LCE compositions with a fixed Ag loading amount of 1.5 wt\% and different Cu loading: 0.75 wt\% (red, circles), 1.5 wt\% (blue, squares), 3 wt\% (green, stars), and 6 wt\% (gray, triangles). }
\label{fig:photothermal_effect}
\end{figure}

With the reduced actuation temperature and improved actuation stress, we next focus on enhancing the photothermal response of the LCE to achieve the required actuation temperature under lower irradiance. Incorporating photothermal agents is a common strategy to improve light absorption and photothermal heating. However, increasing the concentration of these agents presents a trade-off: while it enhances heating, it can also disrupt molecular organization within the LCE network and reduce actuation stress.
We address this through a systematic exploration of photothermal-agent type and agent-to-LCE mass ratio based on the optimized material formulation (5-2). We evaluate the temperature rise of different photothermal agents under illumination from a solar simulator (Model 10500, Abet Technologies, Milford, CT). 

Dark photothermal agents such as candle soot interfere with effective crosslinking, particularly in thicker samples. We therefore focus on optimizing the loading of metallic microparticles, specifically silver (Ag) and copper (Cu). We systematically vary the Cu loading while keeping the Ag content fixed at 1.5 wt\% to evaluate its effect on photothermal performance. Formulations containing 0.75–3 wt\% Cu generate sufficient photothermal heating to trigger snap-through under 1 Sun illumination when bonded to Ecoflex 00-50 elastomer. Using a thermal camera, we track the temperature evolution of individual LCE strips and the assembled device  (Fig.~\ref{fig:photothermal_effect}a) and correlate it with the deformation of the composite beam during heating-induced snap-through (Fig.~\ref{fig:photothermal_effect}b).

We find that increasing copper loadings typically increases the heating rate (Fig. \ref{fig:photothermal_effect}c). However, the photothermal agents act as impurities and may hinder the reorganization of the LCE network; consequently, increasing the loading of Ag and Cu reduces the actuation stress (Fig. \ref{fig:photothermal_effect}d). To balance these effects, we identify 1.5 wt\% Ag and 1.5 wt\% Cu as the optimal composition, producing the fastest photothermal heating and mechanical actuation at both the material and device levels (Fig. \ref{fig:photothermal_effect}e). We verify the snapping behavior by tracking changes in arc height of the assembled device during snapping (Fig. \ref{fig:photothermal_effect}b). The optimized formulation exhibits the earliest snap-through under illumination with 1 Sun, as shown by the time evolutions of height (Fig. \ref{fig:photothermal_effect}e) and temperature rise (Fig. \ref{fig:photothermal_effect}f).


\subsection*{Tuning snapping energy barrier}
\begin{figure}
\centering
\includegraphics[width=\textwidth]{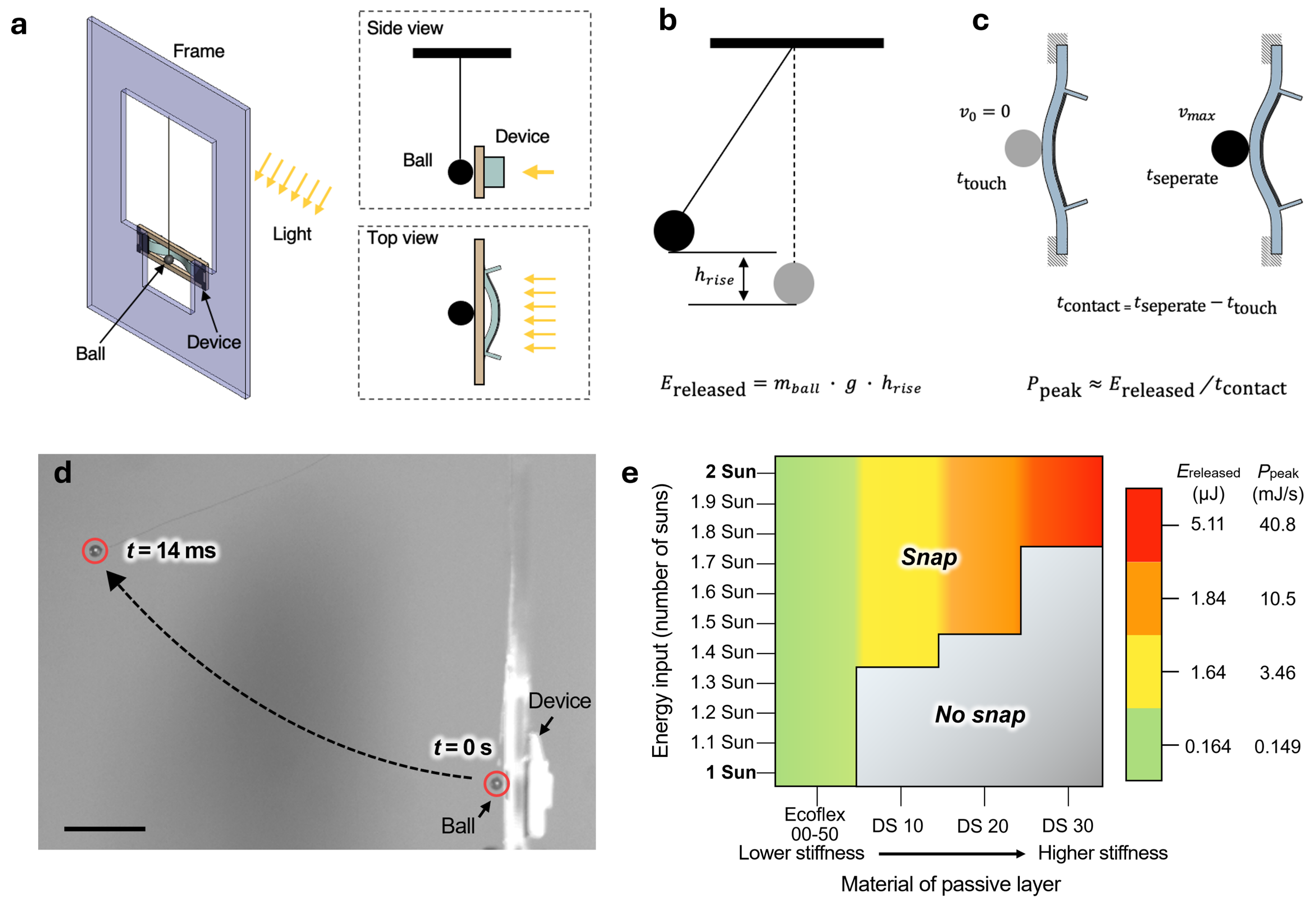}
   \caption{\textbf{Experimental setup and quantification of snap-through energy.} (a) Schematic of the experimental setup shown in side and top views. (b) The released steel ball reaches a maximum height, $h_\mathrm{rise}$, allowing the released energy to be determined from the corresponding increase in potential energy against gravity. (c) Measurement of the contact time during snap-through. (d) Representative side-view montage of the snap-through process. Dragon Skin 30 served as the passive layer. The scale bar is 10 mm. (e) Released energy and peak power for samples with different silicone substrates. The estimated Young's moduli are 0.12, 0.18, 0.36, and 0.52 MPa for Ecoflex 00-50 and Dragon Skin 10, 20, and 30, respectively, based on public data from \citet{Marechal2020}.}
\label{fig:energy_release}
\end{figure}

Matching the energy barrier of the monostable elastomer to the mechanical output of the LCE allows low-intensity illumination to be converted into cyclic, high-power motion. We experimentally quantify the energy released by the composite beam during snap-through~\cite{yan2025self}. Because the combined effects of material composition, geometry, and multiphysics interactions make accurate analytical or numerical predictions challenging, we measure the released energy using a ball-launch experiment. Specifically, we measure the maximum height reached by a steel ball launched during snap-through (Fig.~\ref{fig:energy_release}a). The increase in the ball’s gravitational potential energy, $E_\mathrm{release}=mgh_\mathrm{rise}$, provides a lower-bound estimate of the transferred mechanical energy (Fig.~\ref{fig:energy_release}b). Together with the measured contact time between the beam and ball, this energy was used to estimate the peak mechanical power output (Fig.~\ref{fig:energy_release}c). A representative launch sequence is shown in Fig.~\ref{fig:energy_release}d.

Varying the elastomer stiffness reveals a tradeoff between the actuation threshold and mechanical output (Fig.~\ref{fig:energy_release}e). Softer elastomers require less LCE-generated force to trigger snap-through, enabling operation at lower irradiance but storing less elastic energy and therefore producing lower released energy and peak power. Stiffer elastomers store and release more energy but impose a higher energy barrier, requiring greater LCE-generated stress and eventually preventing snap-through when the elastomer becomes too stiff. The maximum mechanical output is therefore achieved by matching the structural instability threshold to the available actuator stress.

Passive-layer stiffness thus governs both the triggering threshold and the releasable elastic energy, providing an important design parameter for the high-performance jumpers demonstrated in later sections. We compare the peak mass-specific mechanical power of our system with previously reported photothermally driven actuator systems in Fig.~\ref{fig:overview}d. The comparison shows that our system occupies a distinct regime of high specific mechanical power under solar-level irradiance. This high peak specific mechanical power can enable rapid, mechanically demanding actions, allowing robotic functions, like rapid shape change,
jumping, and effective propulsion.



\subsection*{Self-sustained robotic locomotion}

We next demonstrate how the cyclic snap-through and snap-back motions of the composite beam can be used to power self-sustained robotic systems. Its simple architecture allows it to be readily integrated with other functional materials and robotic components, enabling autonomous operation under solar-level (1 Sun) illumination. We illustrate this versatility with several robotic platforms, including a swimmer, a crawler, and a jumper.
\begin{figure}[htb!]
\centering
\includegraphics[width=0.8\textwidth]
{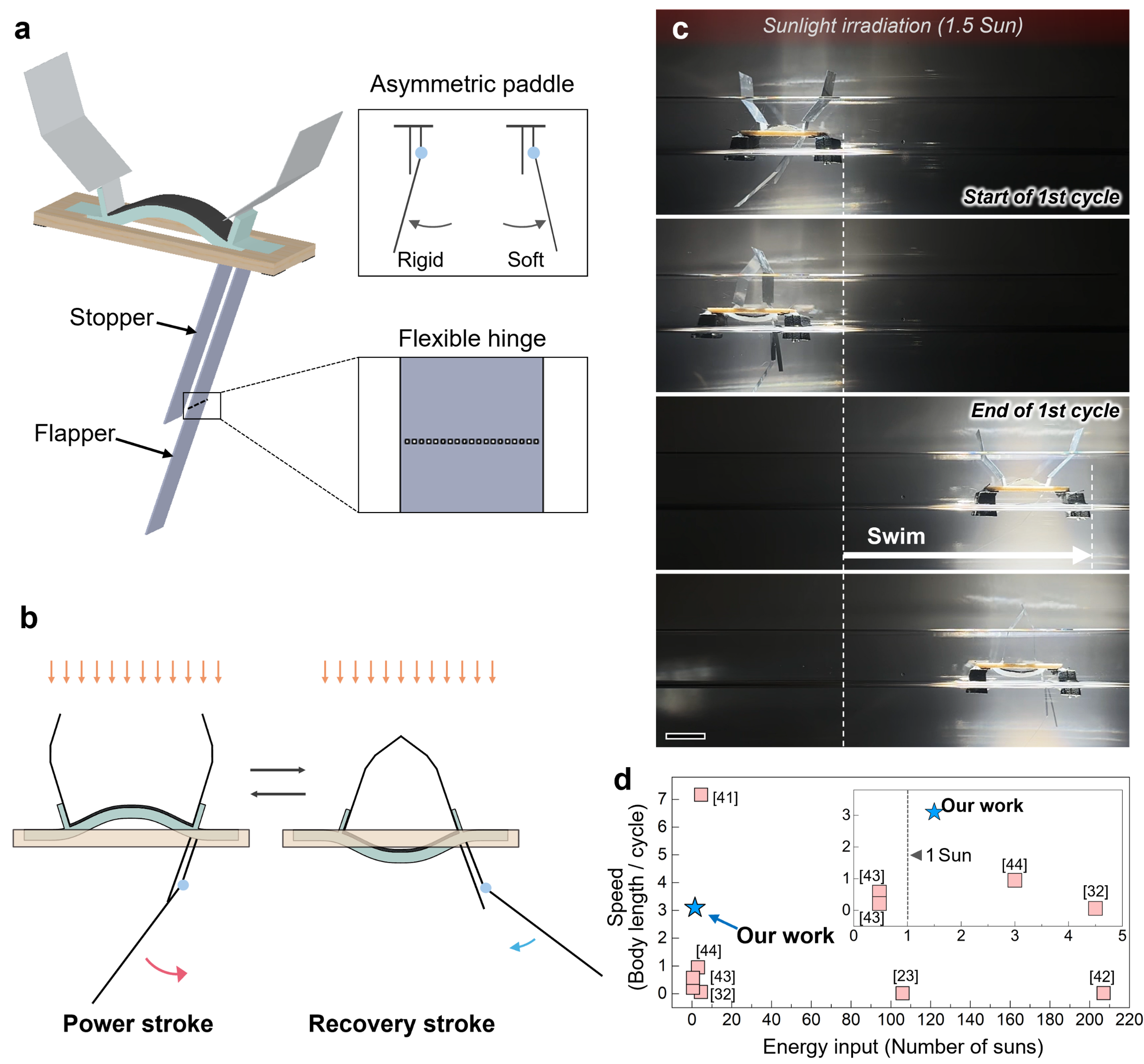}

  \caption{\textbf{Sunlight-powered swimmer.}
(\textbf{a}) Schematic of the sunlight-driven swimming robot. The asymmetric paddle consists of a shorter stopper and a longer flapper with a flexible hinge. The asymmetric paddle design provides direction-dependent bending stiffness, converting snap-through and snap-back motions into directional propulsion underwater. 
(\textbf{b}) Simplified locomotion mechanism of the swimmer. Under illumination, the composite beam bends downward and drives the paddle through a recovery stroke, during which the flapper is soft and bends at the flexible hinge, substantially reducing the backward propulsive force. When the composite beam snaps back, the flapper is constrained by the stopper and behaves as a rigid paddle, generating a large propulsive force forward. The difference between the power and recovery strokes produces net directional locomotion.
(\textbf{c}) Snapshots showing a swimming cycle under 1.5 Suns, including snap-through, snap-back, and the resulting net displacement.
 (\textbf{d}) Comparison of swimming speed and input light irradiance with previously reported light-powered swimmers and related locomotors.}
\label{fig:demo_swimmer}
\end{figure}


\subsubsection*{Swimming of an untethered swimmer on water}

Directional locomotion requires symmetry breaking. We therefore design a paddle with asymmetric stiffness in opposite directions to generate net directional propulsion. The paddle consists of a mechanical stopper and a flapper, with a flexible hinge connecting the two parts of the flapper (Fig. \ref{fig:demo_swimmer}a). The hinge is positioned above the edge of the stopper, allowing the flapper to bend freely in the clockwise direction with low stiffness. In contrast, bending in the opposite direction is constrained by the stopper, resulting in much higher effective stiffness. This stiffness contrast creates asymmetric hydrodynamic drag between the two bending directions, which is desirable, as it 
increases the net propulsive force. The paddle is attached underneath the curved beam and positioned near the base of the shutter to maximize its rotational angle, thereby generating a larger paddle excursion.
 
Upon illumination, the composite beam bends downward and snaps through, executing a recovery stroke and generating only a small propulsive force due to the compliance of the flexible hinge. In the meantime, it closes the shutter to block the illumination. As the system cools and the composite beam snaps back, it generates a power stroke that propels the swimmer rightward (Fig. \ref{fig:demo_swimmer}b). Over each cycle, the asymmetric propulsion between the power and recovery strokes produces a net rightward displacement, enabling the swimmer to sustain directional locomotion under constant illumination. 

\begin{figure}[htb!]
\centering
\includegraphics[width=0.7\textwidth]{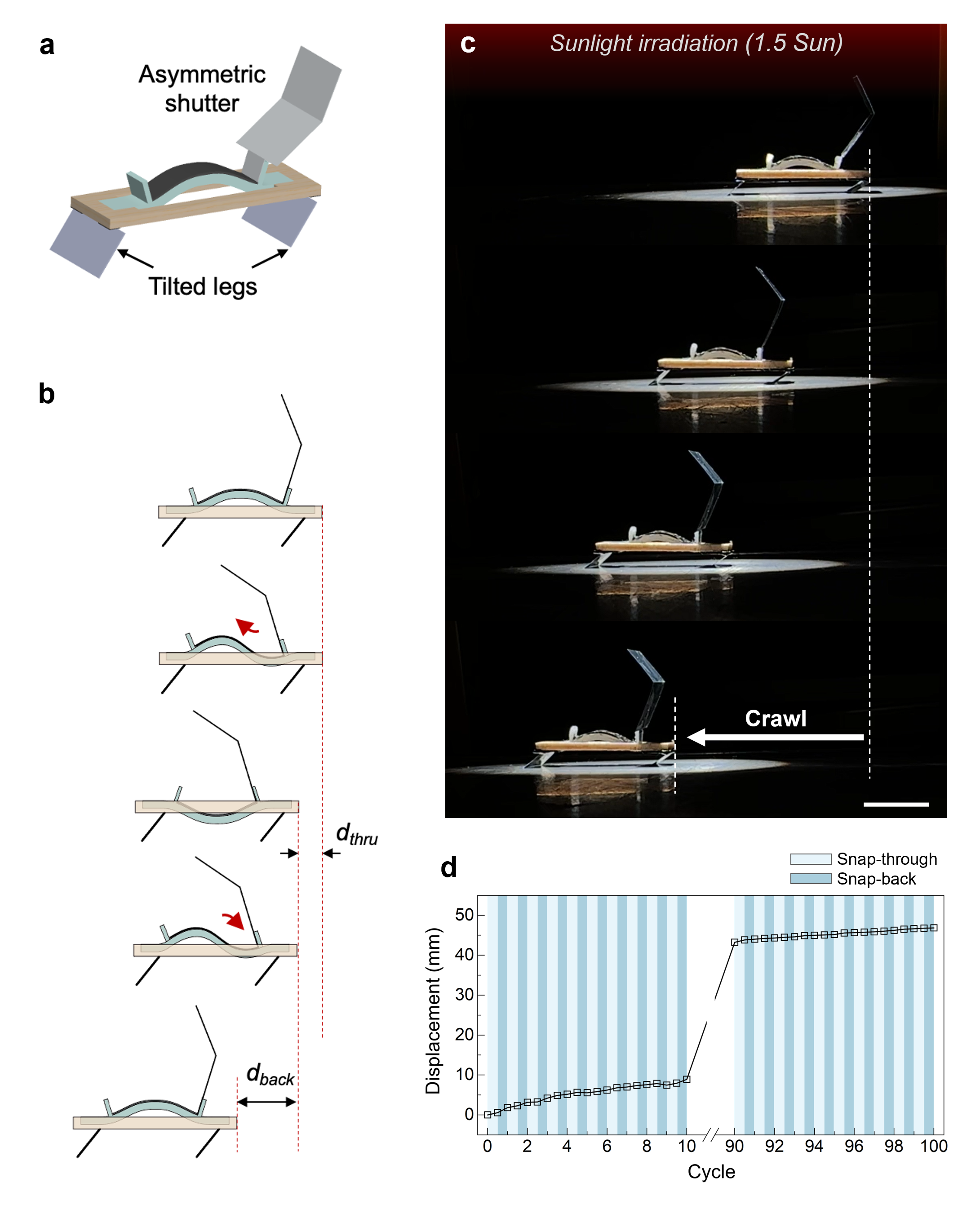}
   \caption{\textbf{Sunlight-powered crawler.} (\textbf{a}) Schematic of the sunlight-powered crawling robot.  
   (\textbf{b}) Simplified locomotion mechanism of the crawler over successive actuation cycles. The robot harnesses an asymmetric snap-back transition, during which the composite beam adopts an S-shaped configuration that drives rightward motion. The tilted leg provides direction-dependent friction, further promoting directional locomotion.
   (\textbf{c}) Under 1.5 Suns, self-sustained oscillation of the composite beam drives the crawler to move along the surface (Supplementary Movie S4). Scale bar: 15 mm.
   (\textbf{d}) Displacement of the crawler over time, showing stepwise locomotion driven by repeated oscillation cycles.
    }
\label{fig:Crawler}
\end{figure}
The systematic exploration and optimization, the swimmer prototype can continuously propel itself under an illumination of only 1.5 Suns (Fig. \ref{fig:demo_swimmer}c). After a brief resting period, the swimmer is powered by constant illumination and travels 3.1~body lengths/cycle (Fig. \ref{fig:demo_swimmer}d and Supplementary Movie S3). This represents one of the fastest locomotion speeds among sources surveyed \cite{yin2021visible, shi2025air, cezan2024autonomous, zhao_soft_2019, yang2022bioinspired, li2021light} for self-sustained light-powered swimmers, particularly considering the low energy input required for continuous operation (Supplementary Table S3).

The low-cost, nonmagnetic, and electronics-free architecture could be advantageous for robotic operation in remote or extreme environments, including those with strong magnetic fields or high radiation levels. The sustained water-surface locomotion also offers opportunities for environmental remediation; for example, integrating the swimmer with oil-absorbing materials could enable continuous collection of floating oil while traversing contaminated water surfaces.

\subsubsection*{Crawler: directional locomotion}
Compared with swimming, locomotion on solid surfaces is more challenging because of increased friction with the substrate~\cite{yan2022crawling}. To achieve untethered crawling, we use the rapid snap-through and snap-back motions of the composite beam to drive locomotion. To reduce friction from the ground, two tilted legs were attached to the base at an angle of 45$^\circ$ relative to the ground (Fig. \ref{fig:Crawler}a). We remove one shutter to introduce mass asymmetry and promote directional locomotion. During each actuation cycle, the compliant LCE–elastomer composite undergoes asymmetric snap-through and snap-back transitions, following the same principle as Ref.~\cite{yan2025self} (Fig. \ref{fig:Crawler}b). 

The simplified crawling mechanism of the robot can be described in two consecutive steps. First, the beam snaps through, transferring its kinetic energy to the robot as it rapidly transitions through an S-shaped configuration and propelling the robot leftward by a distance $d_{\mathrm{thru}}$. Second, the reaction force generated as the composite beam snaps back through the reverse S-shaped transition drives the crawler farther forward by $d_{\mathrm{back}}$. This transient, impact-induced actuation arising from snap-through and snap-back has been previously reported~\cite{yan2025self}. The robot therefore advances by a total distance $d=d_{\mathrm{thru}}+d_{\mathrm{back}}$ during each actuation cycle. Through repeated cycles, the robot crawls continuously and travels one body length under an illumination intensity of 1.5 Suns (Fig.~\ref{fig:Crawler}c and d).

\subsubsection*{Autonomous, self-sustained jumper}

\begin{figure}[htb!]
\centering
\includegraphics[trim= 0in 0cm 0in 0.1cm, clip=true, width=\textwidth]{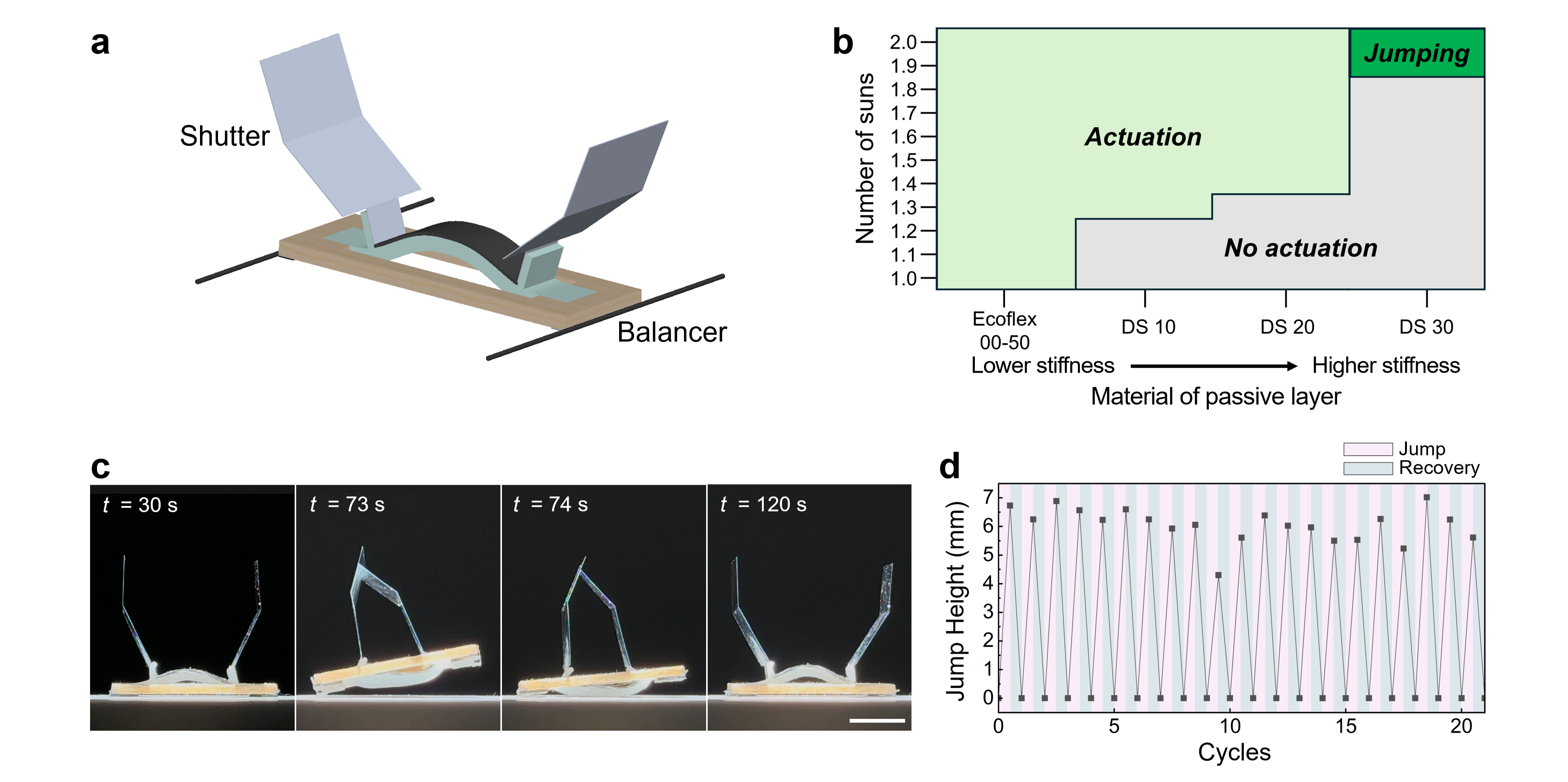}
   \caption{\textbf{Sunlight-powered jumper.} (\textbf{a}) Schematic of the sunlight-powered jumping robot. The robot incorporates shutters and balancers to enable repeated jumping. (\textbf{b}) Effect of passive-layer stiffness on robot motion. Actuation refers to snap-through without takeoff, which occurs under lower sunlight intensity or with softer elastomers. 
   Compared with the softer DS10, DS20, and Ecoflex 00-50, the stiffer DS30 passive layer allows for greater elastic energy storage and release during snap-through, sufficient to drive jumping. The estimated Young's moduli for Ecoflex 00-50 and Dragon Skin 10, 20, and 30 are 0.12, 0.18, 0.36, and 0.52 MPa respectively, based on public data from \citet{Marechal2020}.
   (\textbf{c}) Snapshots showing jumping under 2 Suns, driven by snap-through. The scale bar is 10 mm. (\textbf{d}) Jumping height over actuation cycles, showing reproducibility and no decrease in jumping height after more than 20 cycles.}
\label{fig:jumping_demo}
\end{figure}

The rapid release of energy from the composite beam can also generate impulsive motion, which is challenging to achieve in self-sustained systems. We explore this capability using a self-sustained jumping robot (Fig.~\ref{fig:jumping_demo}a). Jumping requires an energy barrier that is low enough to be overcome by the LCE actuation but high enough to store sufficient elastic energy for propulsion. With a relatively low energy barrier, the composite beam undergoes snap-through without generating sufficient energy for jumping (Fig.~\ref{fig:jumping_demo}b). Using a stiffer elastomer (Dragon Skin 30) increases the energy barrier, allowing more elastic energy to accumulate before snap-through. This enables jumping under approximately 1.9 Suns, substantially lower than the 6 Suns required in our previous work~\cite{yan2025self}. To achieve reversible jumping, we add two shutters to enable cyclic jumping and two balancers at the bottom of the frame to prevent tipping. The robot achieves more than 20 consecutive jumps without a measurable decrease in jump height throughout all cycles (Fig.~\ref{fig:jumping_demo}c and d; Supplementary Movie S5). Although autonomous jumping has been achieved previously, our design enables repeated jumping at solar intensities below 2 Suns.

\section*{Discussion}\label{sec:discussion}

In this work, we demonstrate \emph{SunLightBots}, a class of self-sustained soft robots that transform continuously harvested solar energy into cyclic, high-power mechanical motion. The central advance is integrating energy accumulation, rapid release, and autonomous resetting to sustain cyclic, high-power motion under continuous low-power solar input.  This is achieved through a self-resetting snapping (SRS) mechanism that couples liquid crystal elastomers with snap-through instability to store solar energy delivered at low power, release it in rapid, high-power motion, and reset through embedded photothermal–mechanical feedback. Under illumination, the LCE gradually generates stress that is transferred to an elastomeric structure, where it is stored as elastic energy and rapidly released through snap-through. The total energy is not amplified; instead, its release is concentrated in time to amplify mechanical power. By coordinating material transduction, structural energy storage, and photothermal--mechanical feedback, the SRS mechanism operates near 1 Sun, achieves a peak mass-specific mechanical power of 89~W/kg, and autonomously resets under continuous illumination. 

\emph{SunLightBots} extend self-sustained photomechanical actuation toward power-amplified robotic motion. The swimming, crawling, and jumping demonstrations show that the same mechanism can support sustained aquatic and terrestrial locomotion as well as impulsive movement. The swimmer advances more than three body lengths per cycle, although its average speed remains limited by the slower thermal loading and resetting stages. This result illustrates the distinction between peak and average power: snap-through provides rapid mechanical output, whereas the overall cycle rate remains governed by solar-energy harvesting and heat transfer.
Further improvements in power, efficiency, and cycle rate will require coupled optimization of the active material and unstable structure. At the material level, broader spectral absorption, lower transition temperatures, greater actuation stress, and improved cycling stability could increase the mechanical work generated per unit illuminated area~\cite{yang2026four,sol20234d}. At the structural level, optimizing geometry, modulus, interfacial stress transfer, and instability thresholds could increase power output while preserving reliable triggering and reset~\cite{zhang20263d}. 

Our robotic platforms illustrate the versatility of the SRS mechanism in enabling autonomous, self-sustained locomotion, including swimming, crawling, and jumping. Beyond locomotion, the same mechanism can be used for sensing and control. As a proof of concept, we develop a sunlight-triggered switch that closes an electrical circuit to illuminate a light-emitting diode (Supplementary Fig. S5; Supplementary Movie S6). This example highlights the potential of the SRS mechanism for light-responsive triggering without electronic sensing or control components, as well as its integration into more complex robotic systems.

To achieve more sophisticated robotic autonomy, further advances are necessary, building on the actuation-level autonomy established here. Light-orientation mechanisms could maintain favorable exposure and enable phototactic navigation~\cite{xu2026humidity}. Alternative instabilities and mechanically coupled components could further generate twisting, rotation, and sequential responses for more coordinated locomotion~\cite{zhao2022twisting,zhai2023desktop}. Integrating low-power sensing, stimulus-responsive control, and communication could allow the robots to detect environmental conditions, adapt their behavior, and relay collected information~\cite{zheng2024liquid}. These capabilities must be co-designed within the limited harvested-energy budget, but could ultimately transform self-sustained actuation into autonomous navigation and task execution.

Real-world deployment will additionally require long-term mechanical durability and robust operation under changes in irradiance, angle of incidence, temperature, weather, and environmental loading. Environmentally benign or biodegradable active materials could enable temporary deployment without robot retrieval~\cite{guo2026cellulose}. 

\section{Materials and Methods}\label{sec:methods}

\subsection{Materials}

Reactive monomers RM257 (1,4-bis[4-(3-acryloyloxypropyloxy) benzoyloxy]-2-methylbenzene) and RM82 (1,4-bis[4-(6-acryloyloxyhexyloxy)benzoyloxy]-2-methylbenzene) were purchased from Daken Chemical (Zhengzhou, Henan, China). EDDT (2,2'-(ethylenedioxy)diethanethiol), TATATO (1,3,5-triallyl-1,3,5-triazine-2,4,6(1H,3H,5H)-trione),  dodecylamine, butylated hydroxytoluene (BHT), PEGDA 250 (poly(ethylene glycol) diacrylate, Mn = 250), silver microparticles (powder, 2-3.5 $\mu$m), {copper microparticles (average particle size = 10 $\mu$m) } and triethylamine,  were purchased from Sigma-Aldrich (St. Louis, MO). 4-(4-aminophenyl)butyric acid was purchased from AmBeed (Buffalo Grove, IL). Photoinitiator Irgacure 651 (2,2-dimethoxy-2-phenylacetophenone, I-651) was purchased from MedKoo BioSciences (Durham, NC). All chemicals were used without further modification unless otherwise stated. Candle soot (CS) was obtained from the incomplete combustion of a paraffin candle. An aluminum foil was positioned above the candle flame to collect soot deposits, and the accumulated CS was subsequently removed from the foil surface by gentle scraping.  Water was dispensed from a Milli-Q system with a resistivity 
 = 18.2 M$\Omega\cdot$cm at 25$^\circ$C. 

\subsection{Synthesis of LCE}

LCEs were synthesized via either amine- or thiol-based click reactions using the formulations listed in Supplementary Table S2. For amine-based LCEs, the LC monomer, chain extender, and I-651 were combined in a 20 mL scintillation vial. For thiol-based LCEs, the LC monomer, chain extender, crosslinker, BHT, and I-651 were combined in a 20 mL scintillation vial. In both cases, the mixture was heated above $T\textsubscript{NI}$ and vortexed until homogeneous, followed by thermal oligomerization at 150 $^\circ$C. Amine-based formulations were oligomerized for 15 h, whereas thiol-based formulations were oligomerized for 3.5 h. After oligomerization, the resulting oligomer was cast into a glass mold (10 mm $\times$ 30 mm) and partially crosslinked under 365 nm UV light (intensity = 1.21 mW cm$^{-2}$ ). The partially crosslinked film was then uniaxially stretched to 100\% strain using a mechanical stretching apparatus, corresponding to twice its original length. While maintained under strain, the LCE film was fully crosslinked under UV light for 3 h. To promote uniform curing, the sample was flipped every 5 min during the first 30 min of irradiation and every 30 min thereafter until crosslinking was complete.

\subsection{Thermomechanical characterization}

Thermomechanical characterization was performed using a dynamic mechanical analyzer (DHR 30, TA Instruments). For isometric stress measurements, aligned LCE films were cut into rectangular strips and held at a fixed length while the temperature was increased from 25 to 100 $^\circ$C at a rate of 3 $^\circ$C/min. 

Actuation strain as a function of temperature was measured using an optical microscope equipped with a Linkam thermal stage (The McCrone Group, Westmont, IL). LCE films were first heated to 100 $^\circ$C and then cooled to erase thermal history. The films were subsequently subjected to heating–cooling cycles between 25 and 65 $^\circ$C at a rate of 3 $^\circ$C min$^{-1}$, during which dimensional changes were recorded to obtain strain–temperature profiles.

\subsection{Photothermal characterization}

The light irradiance of the solar simulator was adjusted by varying the distance between the light source and the sample. The corresponding calibration curve, measured independently using a light meter, is shown in Supplementary Fig. S4.

Metal microparticles (silver and copper) were incorporated into the LCE precursor as photothermal absorbers to enhance solar absorption and temperature rise. The rise in temperature was characterized by a thermal camera (Thermal Master P3, Yantai, Shandong, China). 

\subsection{Fabrication of the LCE-elastomer composite} 

The LCE was combined with a curved elastomer beam to form a composite to introduce mechanical instability. The curved beam is obtained by mirroring and then extruding the height profile~\cite{yan2025self} to a certain width, $w$:

\begin{equation}
  h(x) = \begin{cases}
    (H-d) [1-10(\frac{x}{L})^3+15(\frac{x}{L})^4-6(\frac{x}{L})^5], & 0 < x \leq L,\\
    0, & L < x \leq L+s.
  \end{cases}
\label{eq:beamProfile}
\end{equation}
\\
\noindent where $H$ is the overall height, $d$ is the nominal thickness, and $L$ is the span of the arc. $s$ denotes the length of the flat portion added at the base of the beam to facilitate the enforcement of clamped-clamped boundary conditions when integrated into a frame. The final beam geometry is obtained by offsetting the height profile $h(x)$ along its surface normal by a distance $d$ to account for the uniform thickness.

A photocurable mold was printed using a 3D printer (Form 4, Formlabs). Dragon Skin 10, 20, 30 (Macungie, PA) and Ecoflex 00-50 (Legal, Alberta, Canada) were cast into the molds, degassed under vacuum, and cured overnight, followed by additional curing in the oven at 65 $^\circ$C. Afterward, the excess materials were trimmed with a razor blade and then released from the mold. The LCE and silicone were bonded together using a heat-resistant adhesive (500°F RTV High Heat Silicone Sealant, Rutland). 

\subsection{Design of the shutters} 

Autonomous cyclic motion requires a shutter design that can accommodate the full range of motion of the LCE–elastomer composite. We optimized the shutter geometry using 2D simulations to ensure that it closes completely when the composite is actuated downward to its fully deformed state (Supplementary Movie S2). This complete closure shields the LCE from sunlight, thereby allowing the LCE to cool, recover, and reverse its motion, enabling autonomous cyclic actuation.

\subsection{Energy release quantification}

The energy released during snap-through was quantified by impacting the composites with a hollow steel ball weighing 16.5 mg. The process was captured using two high-speed cameras (AX100, Photron, Tokyo, Japan) that simultaneously recorded side and top views of the experiment. The contact time and rebound height were extracted from the high-speed recordings (40,000 frames per second) and used to calculate the energy transferred from the composite to the ball during the snap-through event (Supplementary Movie S1).

\subsection{Fabrication of microrobots}
A balsa wood frame (1.5 mm thick, Qj-solar) was fabricated by laser cutting (VLS3.50, Universal Laser Systems) following the design (Fig. S6). Two carbon fiber substrates were bonded to the bottom of the balsa wood frame using Loctite super glue (Rocky Hill, CT).  
The LCE-elastomer composite was then attached to the carbon-fiber-supported balsa wood frame using Loctite SI 5699 Gray adhesive (Henkel AG \& Co. KGaA). The wood frame constrains the composite and enables snap-through behavior, which serves as the primary actuation mechanism for the robots.

For the crawler, starting from the primary device design, two legs and light-blocking shutters were integrated to enable directional crawling. The legs were fabricated from polyester film (0.002" thick, Dura-Lar Film, Grafix, Ohio) and cut into rectangular strips with dimensions of {12 mm × 4.7 mm}. Each strip was manually folded at approximately one-third of its length to form an angle of 45°. The fabricated legs were then attached to the balsa wood substrate using cyanoacrylate adhesive. The legs were fabricated from the same polyester film (0.002 in / 0.05 mm thick, Dura-Lar Film, Grafix, Ohio). Shutters were incorporated into the device to facilitate reversible photothermal actuation of the silicone using superglue.

For the swimmer, the wooden frame was coated with a water-repellent spray to reduce water absorption and was elevated above the water surface by attaching a small piece of bubble wrap underneath the frame. A paddle was bonded directly beneath the hinge using super glue, with the hinge positioned on top of the silicone layer. A compliant hinge was formed by scoring the polyester film with a razor blade and repeatedly folding it. Swimming was recorded in a custom-built water tank under different solar illumination conditions. 

For the jumper, two polyethylene terephthalate shutters (DuraLar, Grafix Plastics, Maple Heights, OH) were affixed using super glue to shade the LCE, allowing it to cool down after photothermal heating and snap-through. Two pieces of carbon fiber, which served as balancers, were attached to the bottom of the wooden frame using super glue. This enables repeatable actuation cycles after landing, even under continuous simulated sunlight, thereby allowing recovery after landing.

\section*{Author contributions}

Y.J.L., W.Y., and Y.L. conceived the concept and designed the project. Y.J.L. fabricated and characterized the LCEs and LCE–elastomer composites. W.Y. created the CAD designs and implemented the 3D printing. Y.J.L., W.Y., M.T.C., and A. T. V. designed, constructed, and tested the demos. Z.D. designed the schematics and diagrams. Y.J.L., W.Y., and Y.L. analyzed the data. Y.L., W.Y., and R.K.B. acquired funding, supervised the research, and provided support for supplies, fabrication, and characterization platforms. All authors contributed to the writing of the manuscript and approved the final version of the manuscript.

\section*{Acknowledgements}

The authors thank the Yale Mechanical and Thermal Analysis Instrumentation Core (MTAIC) for support and assistance in this work. We thank Professor Jaehong Kim for providing the solar simulator, Wentao Zhang and Professor Shu Hu for assistance with calibrating the solar power of the solar simulator, Nikolaos Papachatzis and Professor Madhusudhan Venkadesan for the high-speed camera. Y.L. acknowledges partial support from the National Science Foundation under Grant DMR-2541038. {R.K.B. acknowledges support from the Office of Naval Research (ONR) under grant N00014-24-1-2162.} W.Y. acknowledges support from startup funds provided by the Department of Mechanical and Aerospace Engineering,
University of California, Davis. 

\bibliography{sn-bibliography}
\end{document}